\documentclass{sig-alternate-05-2015}

\usepackage{epstopdf}
\usepackage{graphicx}
\usepackage{amsfonts}
\usepackage{amsmath}
\usepackage[small,tight]{subfigure}
\usepackage{balance}
\usepackage{multirow}
\usepackage{booktabs}
\usepackage{comment}
\usepackage{placeins}
\includecomment{versionfigures}
\usepackage{color}
\usepackage{soul}

\begin{document}

% Copyright
%\setcopyright{acmcopyright}
%\setcopyright{acmlicensed}
%\setcopyright{rightsretained}
%\setcopyright{usgov}
%\setcopyright{usgovmixed}
%\setcopyright{cagov}
%\setcopyright{cagovmixed}

%%%%%%%%%%%%%%%%%%%%%%%%%%%%%%%%%%%%%%%%%%%%%%%%%%%%%%%%%%%%%%%%%%%%%%%%%%%%%%%
%% DOI
%\doi{10.475/123_4}
%
%% ISBN
%\isbn{123-4567-24-567/08/06}
%
%%Conference
%\conferenceinfo{PLDI '13}{June 16--19, 2013, Seattle, WA, USA}
%
%\acmPrice{\$15.00}
%
%%
%% --- Author Metadata here ---
%\conferenceinfo{SIGSPATIAL}{'16 San Francisco, California USA}
%%\CopyrightYear{2007} % Allows default copyright year (20XX) to be over-ridden - IF NEED BE.
%%\crdata{0-12345-67-8/90/01}  % Allows default copyright data (0-89791-88-6/97/05) to be over-ridden - IF NEED BE.
%%%%%%%%%%%%%%%%%%%%%%%%%%%%%%%%%%%%%%%%%%%%%%%%%%%%%%%%%%%%%%%%%%%%%%%%%%%%%%%%%%%%%%%%%%%%
% --- End of Author Metadata ---

\title{Spatio-Temporal Sentiment Hotspot Detection Using Geotagged Photos}

\numberofauthors{1} 
\author{
	%\alignauthor
	Yi Zhu and Shawn Newsam\\
	\affaddr{Electrical Engineering \& Computer Science}\\
	\affaddr{University of California at Merced}\\
	\email{yzhu25,snewsam@ucmerced.edu}
}
\CopyrightYear{2016} 
\setcopyright{acmcopyright}
\conferenceinfo{SIGSPATIAL'16,}{October 31-November 03, 2016, Burlingame, CA, USA}
\isbn{978-1-4503-4589-7/16/10}\acmPrice{\$15.00}
\doi{http://dx.doi.org/10.1145/2996913.2996978}

\maketitle
\begin{abstract}
We perform spatio-temporal analysis of public sentiment using geotagged photo collections. We develop a deep learning-based classifier that predicts the emotion conveyed by an image. This allows us to associate sentiment with place. We perform spatial hotspot detection and show that different emotions have distinct spatial distributions that match expectations. We also perform temporal analysis using the capture time of the photos. Our spatio-temporal hotspot detection correctly identifies emerging concentrations of specific emotions and year-by-year analyses of select locations show there are strong temporal correlations between the predicted emotions and known events.
\end{abstract}



%  Use this command to print the description


%\begin{figure}[htb]
%	\centering
%	\includegraphics[width=1.0\linewidth,trim=0 0 0 0,clip]{figures/firstpage.eps}
%	\vspace{-3ex}
%	\caption{We study the problem of detecting emotions in geotagged images to map public sentiment. Shown are the result of applying our emotion classifier to images from San Francisco. We show that performing this analysis on large collections of images can detect and map public sentiment. This figure is best viewed in color. \vspace{0ex}}
%	\label{fig:firstpage}
%\end{figure}
\vspace{-1ex}
\section{Introduction}
\label{sec:introduction}
Spatio-temporal hotspot detection is an important component of making cities smart, especially for tasks such as monitoring, early warning, resource allocation, and sustainable management. Hotspot analysis is typically conducted by mapping crime rates, monitoring disease outbreaks, locating traffic accidents, etc. In this paper, we focus instead on determining the emotional states of a city's inhabitants as conveyed through their photos as a step towards creating an affect-aware city. 

Emotions play important roles in everyone's daily life. No matter what you do, you will have feelings associated with your activities. Services like Twitter, Facebook, Flickr, or Snapchat are great platforms for people to share their emotional states by posting words/pictures/videos. With geotagged and timestamped social multimedia, we can associate sentiment with geographical locations over time. 
%This motivates our work on spatio-temporal hotspot detection of public emotion.

There exists work on detecting emotions from text such as in Twitter posts \cite{emotionTwitter_2016}, but much less work on using image/video data. The reason is simple, words can deterministically express people's emotions, like ``This is awesome!'', ``I feel blue'', ``So upset''. However, predicting emotions in visual data is much more subtle and difficult. Luckily, the field of computer vision has made great advances recently in high-level image understanding thanks in large part to deep learning. With respect to our problem, large-scale visual datasets for emotion recognition \cite{emotion_image_Peng_2015,Emotion_Large_AAAI16_You,emotion_video_jiang_2014} have been created allowing deep neural networks  to be trained and achieve respectable performance on emotion recognition over the last five years. This has opened the opportunity for work such as ours to exploit these advances for performing spatio-temporal hotspot detection of public emotion using geo-referenced photos.

%This is a hard problem though. Figure \ref{fig:firstpage} shows the predictions made by our emotion classifier for Flickr images from San Francisco. While evaluating whether these labels are correct on a per image basis is not as simple as in an object recognition task, for example, these images do convey the assigned emotions. Further, we believe that performing this analysis on large sets of images, numbering in the millions, will average out the noise associated with the problem and detect reliable signals. 

%Our work exploits the large collections of geotagged photos available at social multimedia sites, the more accurate geotags assigned to these photos, and advanced visual emotion analysis techniques, to assign public sentiment to place. This in turn allows us to perform spatio-temporal hotspot detection.

The major contributions of our work include: 
\textbf{(i)} We conduct the first investigation into sentiment hotspot detection in space and time via geotagged photos. 
\textbf{(ii)} The spatial hotspots for the different emotions have distinct spatial distributions and agree with expectations. 
\textbf{(iii)} Our temporal hotspot analysis is able to detect emerging concentrations of emotions. And, a year-by-year analysis of specific regions finds strong correlations between emotions and temporal events, such as between the level of joy and the success of the San Francisco Giants at AT\&T park, and between the level of disgust and the increase in gentrification in the Mission residential neighborhood.

%\begin{itemize}
%
%	\item We conduct the first investigation into sentiment hotspot detection in space and time via geotagged photos.
%
%	\item The spatial hotspots for the different emotions have distinct spatial distributions and agree with expectations.
%
%	\item Our temporal hotspot analysis is able to detect emerging concentrations of emotions. And, a year-by-year analysis of specific regions finds strong correlations between emotions and temporal events, such as between the level of joy and the success of the San Francisco Giants at AT\&T park, and between the level of disgust and the increase in gentrification in the Mission neighborhood.
%	
%\end{itemize}

%The rest of the paper is organized as follows. After discussing related work in section \ref{sec:related_work}, we describe the details of our emotion classifier and methodology for detecting spatial and spatio-temporal hotspots in section \ref{sec:methodology}. Section \ref{sec:experiments} presents the experiments and results, and the conclusion follows in section \ref{sec:conclusion}.
\vspace{-2ex}
\section{Related Work}
\label{sec:related_work}
\noindent Our work is related to several lines of research.

\noindent \textbf{Geo-Referenced Multimedia} 
The exponential growth of publicly available geo-referenced multimedia has created a range of interesting opportunities to learn about our world. At the intersection of geographic information science and computer vision, large collections of geotagged photos have been used to map world phenomena \cite{CrandallMapWorld09}, classify land use \cite{landuse_yi_sigspatial_2015}, geolocate photos \cite{Haysim2gps08}, recognize and model landmarks \cite{3dModelSnavely08}, perform smart city and urban planning \cite{smartUrban15}, etc.
%and undertake ecological discovery \cite{ZhangEcological12}.
%%% I removed the following from the above since is it not about location:
%%%  analyze sentiment analysis \cite{sentiment_progressive_You_AAAI2015},
Although online photo collections represent a wealth of information, they present challenges due to how noisy and diverse they are. The challenges in using them for geographic discovery include inaccurate location information, uneven spatial distribution, and varying photographer intent. We are mindful of these and recognize they likely temper our results.

%Our work is novel in that it uses a large collection of geotagged photos to map sentiment as conveyed through the photos that ordinary people take. We specifically focus on spatial and spatio-temporal sentiment hotspot detection in an urban area.

\noindent \textbf{Deep Learning}
Deep learning has advanced a number of pattern recognition and machine learning areas including computer vision in which it debuted as deep convolutional neural networks (ConvNets) in 2012 \cite{AlexImageNet12}. Since then, researchers have applied deep ConvNets to a range of vision problems, obtaining state-of-the-art results. Key to ConvNets' performance is their ability to learn high-level or semantic features from the data as opposed to the hand-crafted low- to mid-level features traditionally used in image analysis. This level of image analysis, or understanding, is important for our task since detecting the emotion conveyed in an image is a high-level and abstract task.

\noindent \textbf{Emotion Recognition}
Emotions represent higher intelligence and so being able to recognize them is key to artificial intelligence. For example, real-time emotion recognition during a customer service phone call can lead to a more satisfactory experience; analyzing a Twitter user's emotional state can help detect an emotional crisis; and, a chatting robot who is able to recognize emotions can have better interaction with users. 

Emotions can be conveyed and therefore detected, at least in principle, in various multimedia sources such as text, images, and videos. We develop our own deep learning based system to detect the emotions conveyed in geotagged images. This then allows us to associate sentiment with place.

%\noindent \textbf{Hotspot Analysis}
%Interest in identifying local patterns (hotspots) in spatial data has a long history including the introduction of the popular Moran's I statistic \cite{Moran_1948} in $1948$. Since then, various approaches to finding hotspots have been developed, e.g., spatial and temporal analysis of crime (STAC) \cite{STAC_2006}, space and time scan statistic (SaTScan) \cite{Kulldorff_2001,PULSE_2010_SThotspot_patil,dengue_Naish_2014}, K-means \cite{Grubesic_kmeans_2001}, fuzzy clustering \cite{fuzzy_Grubesic_2006}, Getis-Ord Gi* statistic (Gi*) \cite{ord_getis_1995}, etc.
%Among these approaches, we select the Gi* statistic as our measure to locate spatial patterns which are statistically significant, since  it can provide detailed boundaries for candidate hotspot areas. 

%\begin{figure*}[tb]
%	\centering
%	\includegraphics[width=1.0\linewidth,trim=0 200 0 0,clip]{figures/imageSamples.eps}
%	\vspace{-3ex}
%	\caption{Sample results from applying our image-based emotion classifier to geotagged Flickr images. This figure is best viewed in color. \vspace{0ex}}
%	\label{fig:sampleimage}
%\end{figure*}
\vspace{-2ex}
\section{Methodology}
\label{sec:methodology}
We first describe our approach to detecting the emotion conveyed  by an image. We then describe our spatial and spatio-temporal hotspot detection using the Gi* statistic \cite{ord_getis_1995} and Mann-Kendall test \cite{mann_kendall_test_1945}.
\vspace{-1ex}
\subsection{Emotion Recognition}
As mentioned above, emotion can be conveyed by a number of multimedia sources such as text, audio, image, videos, etc. Visual emotion analysis is appealing since vision, as the richest sense, is arguably the most effective at conveying emotion. 
Existing work on visual emotion analysis can be classified into two approaches, dimensional models \cite{shape_emotion_mm12_Lu} and categorical models \cite{emotion_image_Peng_2015,Emotion_Large_AAAI16_You}. 
We focus on categorical analysis using Ekman's six basic emotions \cite{ekman_6emotions}: \textit{anger, disgust, fear, joy, sadness, and surprise}.
%The dimensional approach is based on a two-dimensional space, valence and arousal (VA) \cite{Russell_VA_model_1980}. Valence describes the emotion on a scale of positive to negative, while arousal indicates the degree of stimulation. The categorical approach, on the other hand, models each emotion as a different class, and can thus be formulated as a classification problem.  
%There are also other variants, like the eight emotion categories considered in the well-known Plutchik's wheel of emotions \cite{Plutchik_8Emotions_97}. 
%Plutchik suggested $8$ primary bipolar emotions: joy versus sadness; anger versus fear; trust versus disgust; and surprise versus anticipation.

Our goal is using geotagged photos for sentiment hotspot detection. The foundation of our approach is assigning each photo one of the six emotions. We therefore design a per-image emotion classifier using ConvNets. This is motivated by the finding of You et al. \cite{Emotion_Large_AAAI16_You} that ConvNets outperform traditional hand-crafted low-level features on most classes in a visual emotion analysis task.
%Our ConvNets based classifier requires a large labeled dataset for training. 
%The training images do not need location information and so we use Emotion6 \cite{emotion_image_Peng_2015} as our training dataset (more on this dataset below). 
%This dataset is not large enough by itself to prevent model overfitting though and so we adopt the common practice of transfer learning in which a ConvNets classifier is first trained on a different but related task for which there is ample training data, and then fine-tuned on the task at hand. 
Specifically, we start with a VGG-16 network that has been pre-trained on ImageNet \cite{imagenet_cvpr09}, and then fine-tune it using the Emotion6 dataset \cite{emotion_image_Peng_2015}.
Once trained, our classifier achieves an average accuracy of $61.95\%$, which is reasonable and performs much better than random guess (which is $16.67\%$).
%The per-class accuracies are shown in Table \ref{tab:emotion6}. 
%Considering the fact that emotion recognition from images is quite an abstract task and one photo may convey a combination of different emotions, the mean accuracy 

%After fine-tuning, we use our trained network to predict the emotions conveyed in the geotagged photos. Given an image, the network outputs a $6$-dim softmax score which represents the probabilities of each emotion class. We take the maximum as the predicted label. 

%We also investigate using the pre-trained VGG-16 network as a feature extractor instead of an end-to-end classifier. These features are fed to a support vector machine (SVM) classifier for which Emotion6 is a sufficiently large training set. This approach turns out to be less effective than using the softmax scores of the fine-tuned VGG-16 network.
\vspace{-1ex}
\subsection{Spatial Hotspot Detection}
Once we have labeled the geotagged images, we can map and start to investigate the spatial distribution of public sentiment. To simplify the analysis, we divide our study area, the city of San Francisco, into a $1000 \times 1000$ grid and assign each image to the closest bin center. The resulting quantization of the image locations does not affect our results since each bin measures less than approximately $12 \times 14$ meters which is finer than the scale of our analysis. All the spatial analysis below is based on the grid instead of the point locations of the photos.

Our data can now be considered a $1000 \times 1000 \times 6$ datacube in which the third dimension is the number of images labeled with a particular emotion. We normalize for the uneven spatial distribution of the images by computing the ratio of each emotion in each bin. That is, for each emotion, we compute a $1000 \times 1000$ grid where each bin is assigned
\begin{equation}
	\text{ratio}_{k}^{e} = \frac{\text{number of photos in bin k of emotion e}}{\text{number of photos in bin k}},
	\label{eq:grid_ratio}
\end{equation}
where $k$ is the spatial index of the bin, and $e$ is the emotion class. Each value in a bin indicates the percentage of a particular emotion evoked at the bin's location. Hence, for each location, the third dimension should sum to $1$.

We use the Getis-Ord Gi* statistic \cite{ord_getis_1995} to find where high and low emotion ratios cluster spatially. Note that, for each emotion, only bins that contain photos are considered and nothing is computed for bins that do not have any photos.

\vspace{-1ex}
\subsection{Spatio-Temporal Hotspot Detection}
Our geotagged photos have timestamps which enables us to perform temporal analysis. These timestamps indicate when the photo was taken. We temporally bin the photos at yearly intervals. Our photos span ten years so we now have a $1000 \times 1000 \times 10 \times 6$ datacube in which each bin is the ratio of images with a particular emotion to the total images for a particular location for a particular year.
This now allows us to perform spatio-temporal hotspot detection.

\noindent \textbf{Global Detection}
We perform spatio-temporal hotspot detection using the emerging hotspot analysis tool\footnote{https://desktop.arcgis.com/en/arcmap/latest/tools/space-time-pattern-mining-toolbox/emerginghotspots.htm} in ArcGIS. We perform this analysis for each emotion separately. First, the Gi* statistic is computed spatially for each year. This is then followed by a Mann-Kendall test \cite{mann_kendall_test_1945} to detect temporal trends at each spatial location. This test essentially looks for correlations between a spatial location's Gi* value and time. The emerging hotspot analysis tool classifies each spatial location into one of 17 categories: new hot (cold) spot, consecutive hot (cold) spot, intensifying hot (cold) spot, persistent hot (cold) spot, diminishing hot (cold) spot, sporadic hot (cold) spot, oscillating hot (cold) spot, historical hot (cold) spot, and no trend detected.

\noindent \textbf{Local Temporal Analysis}
The emerging hotspot analysis tool identifies spatio-temporal hotspots but does not provide detailed information on the year-to-year changes. We therefore perform local analysis at a few locations with the goal of relating the changes to known temporal events. We explore a region's emotional trend over time by computing the emotion ratio for a bounding-box at yearly intervals: 
\begin{equation}
	\text{ratio}_{y}^{e} = \frac{\text{Number of photos locally of emotion e in year y}}{\text{Number of photos locally in year y}}.
	\label{eq:local-ratio}
\end{equation}

\begin{figure*}[!ht]
	\centering
	% \subfigure[Anger]{\includegraphics[width=0.495\linewidth,trim=50 40 50 50,clip]{figures/anger.eps}\label{fig:anger}}\hspace{0pt}
	\subfigure[]{\includegraphics[width=0.32\linewidth,trim=50 40 50 50,clip]{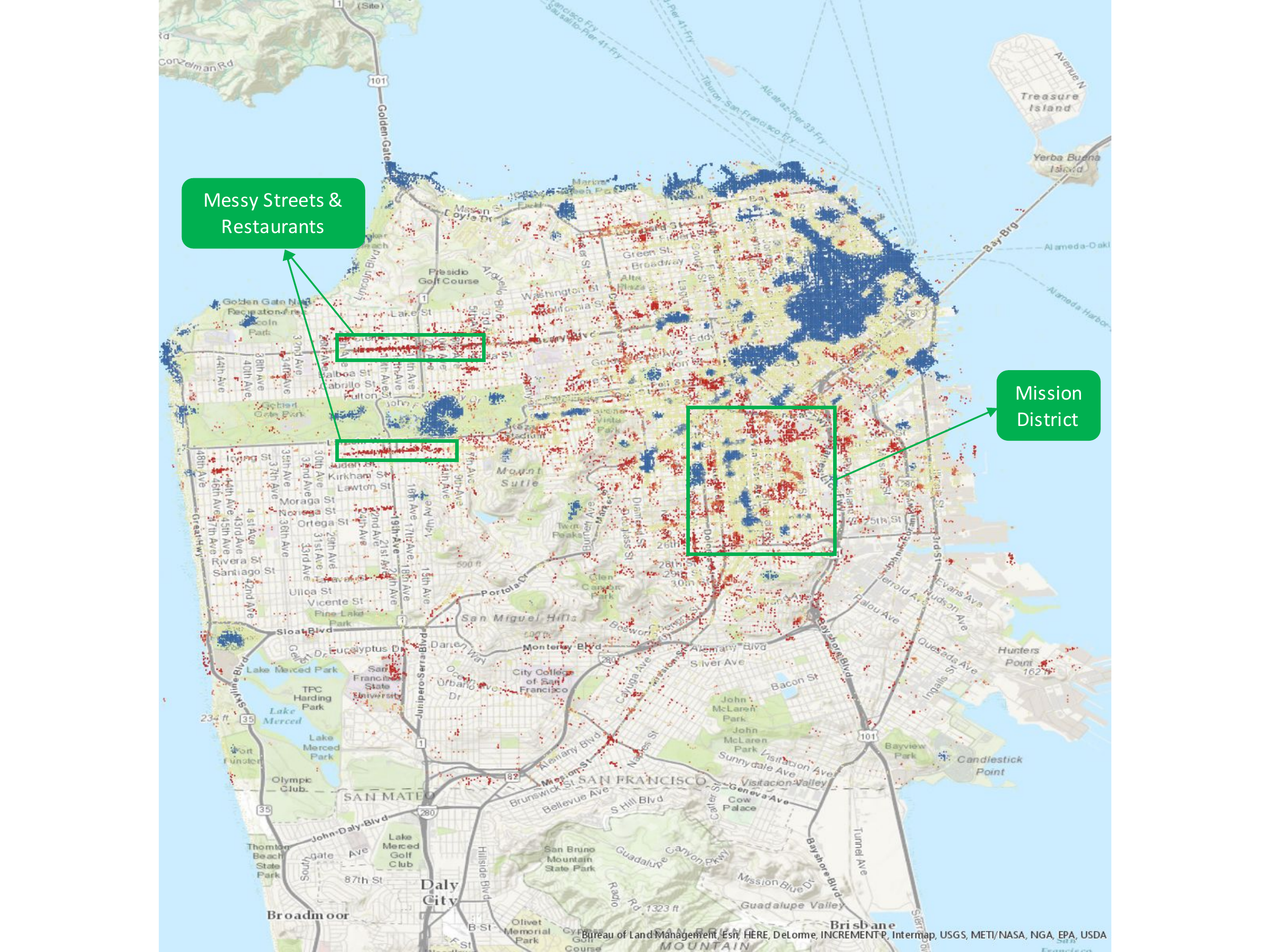}\label{fig:disgust}}
	% \subfigure[Fear]{\includegraphics[width=0.495\linewidth,trim=50 40 50 50,clip]{figures/fear.eps}\label{fig:fear}}\hspace{0pt}
	\subfigure[]{\includegraphics[width=0.32\linewidth,trim=50 40 50 50,clip]{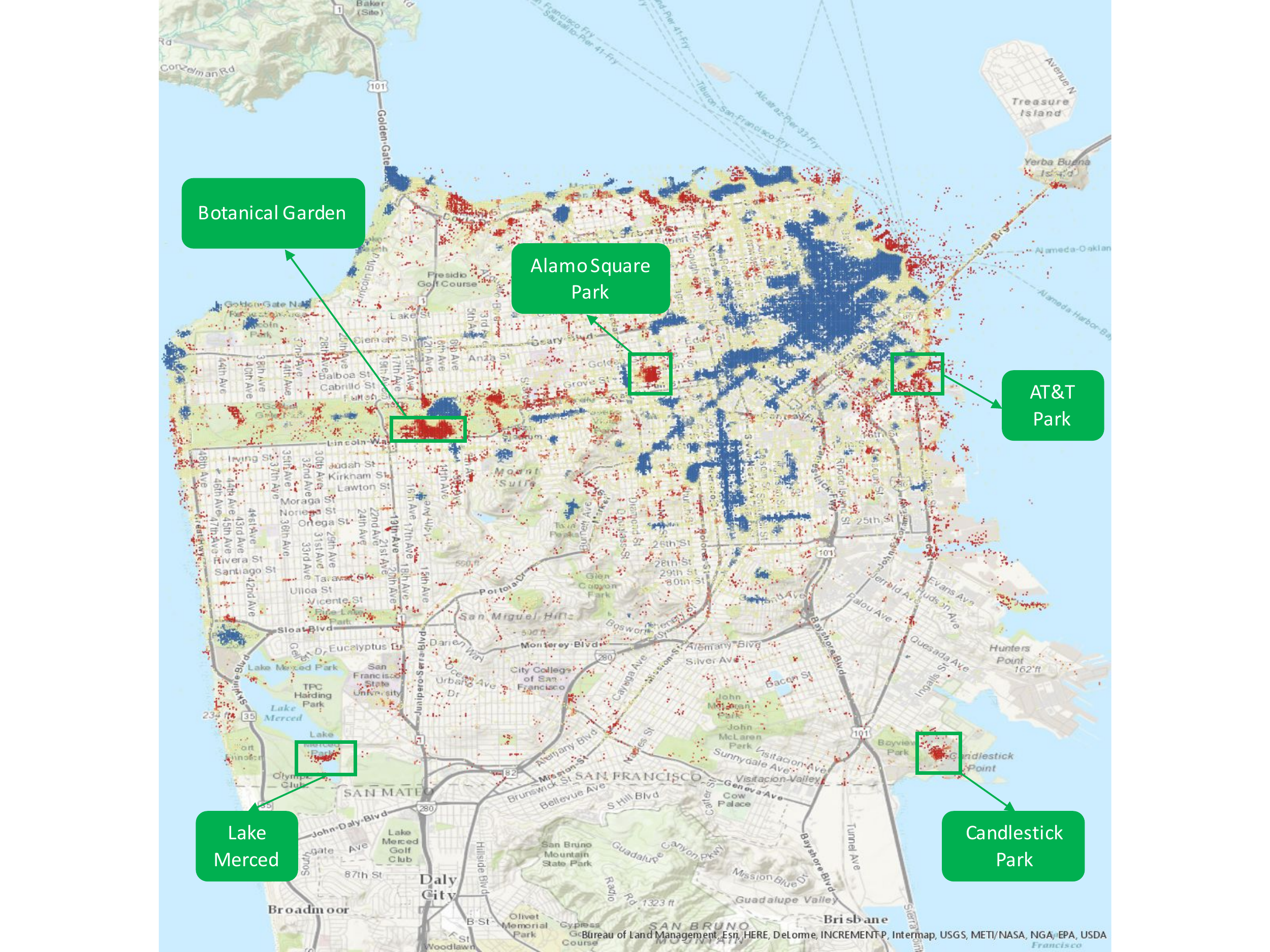}\label{fig:joy}}
	\subfigure[]{\includegraphics[width=0.32\linewidth,trim=0 20 0 0,clip]{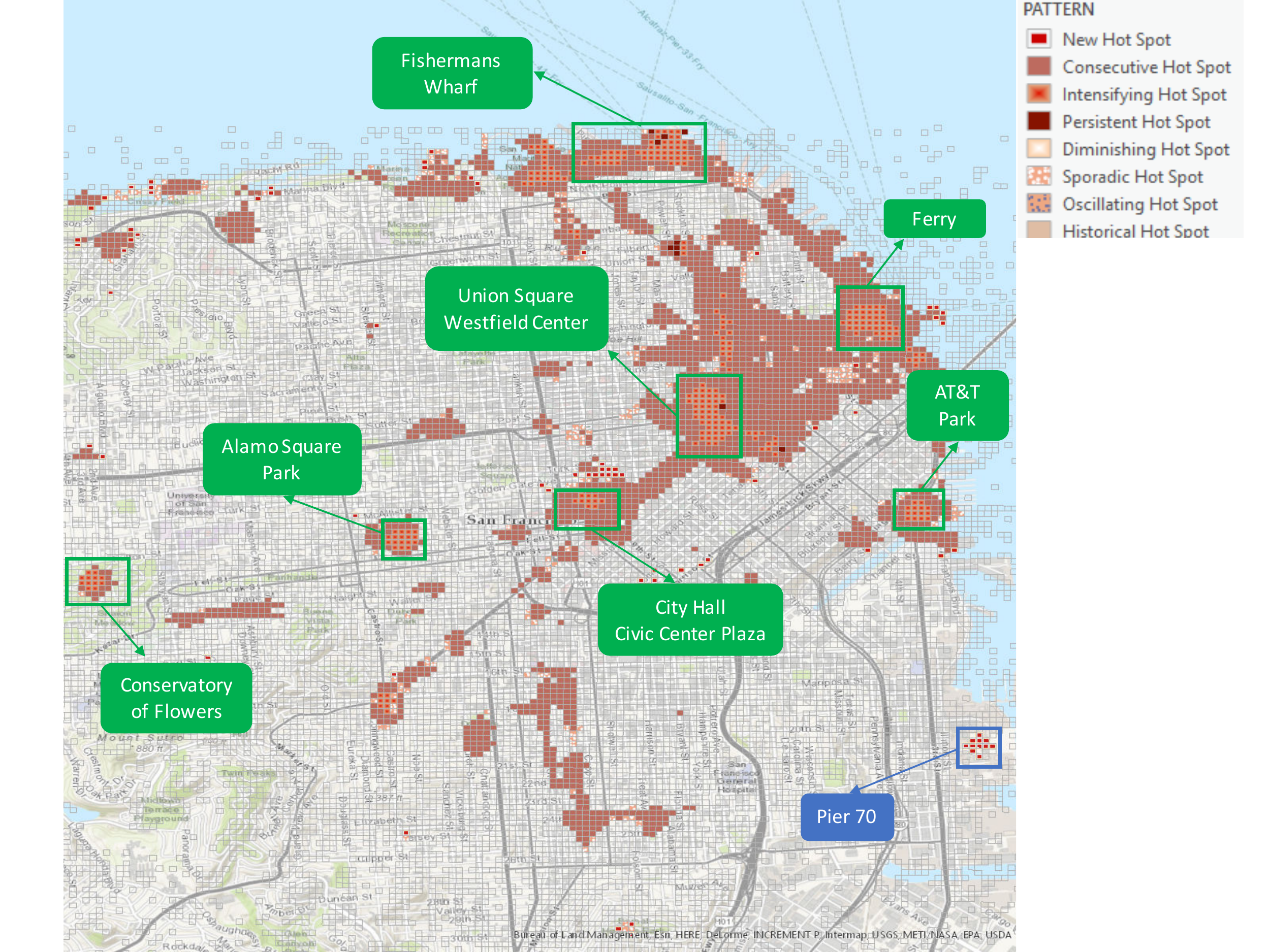}\label{fig:spatialTemporalHotspotDection}}
	% \subfigure[Sadness]{\includegraphics[width=0.495\linewidth,trim=50 40 50 50,clip]{figures/sadness.eps}\label{fig:sadness}}\hspace{0pt}
	% \subfigure[Surprise]{\includegraphics[width=0.495\linewidth,trim=50 40 50 50,clip]{figures/surprise.eps}\label{fig:surprise}}\vspace{1pt}
	\vspace{-3ex}
	\caption{Spatial hotspot detection: (a) disgust; (b) joy. Red, yellow, blue represent hot, not significant and cold spots, respectively. Spatio-temporal hotspot detection: (c) joy. See the text for more details. }
	\label{fig:spatialHotspotDection}
	\vspace{-3ex}
\end{figure*}

\vspace{-2ex}
\section{Experiments and Results}
\label{sec:experiments}
We first describe the training and performance of our emotion classifier. We then present the results of the hotspot detection both in time and space.

%The goal of our experiments is two-fold. First, to obtain a high accuracy emotion classifier. Second, to show that our framework can map and detect hotspots of emotions in large-scale urban areas. Specifically, we first describe our training and testing dataset, followed by the implementation of our emotion classifier. Then we conduct spatial and spatio-temporal hotspot analysis over San Francisco area to make some observations.
\vspace{-1ex}
\subsection{Datasets}
\label{sec:dataset}

\noindent \textbf{Emotion6}
We use the Emotion6 \cite{emotion_image_Peng_2015} image dataset to fine-tune our emotion classifier. This dataset contains 1,980 images evenly divided into six emotion classes: anger, disgust, fear, joy, sadness, and surprise. The images were collected from Flickr by using the class labels and synonyms as search terms. We randomly split the dataset into training and validation subsets in the ratio $8:2$. All images are resized to $256 \times 256$ pixels for input to the classifier.

\begin{figure}[htb]
	\vspace{-2ex}
	\centering
	\subfigure{\includegraphics[width=0.49\linewidth,trim=0 0 0 0,clip]{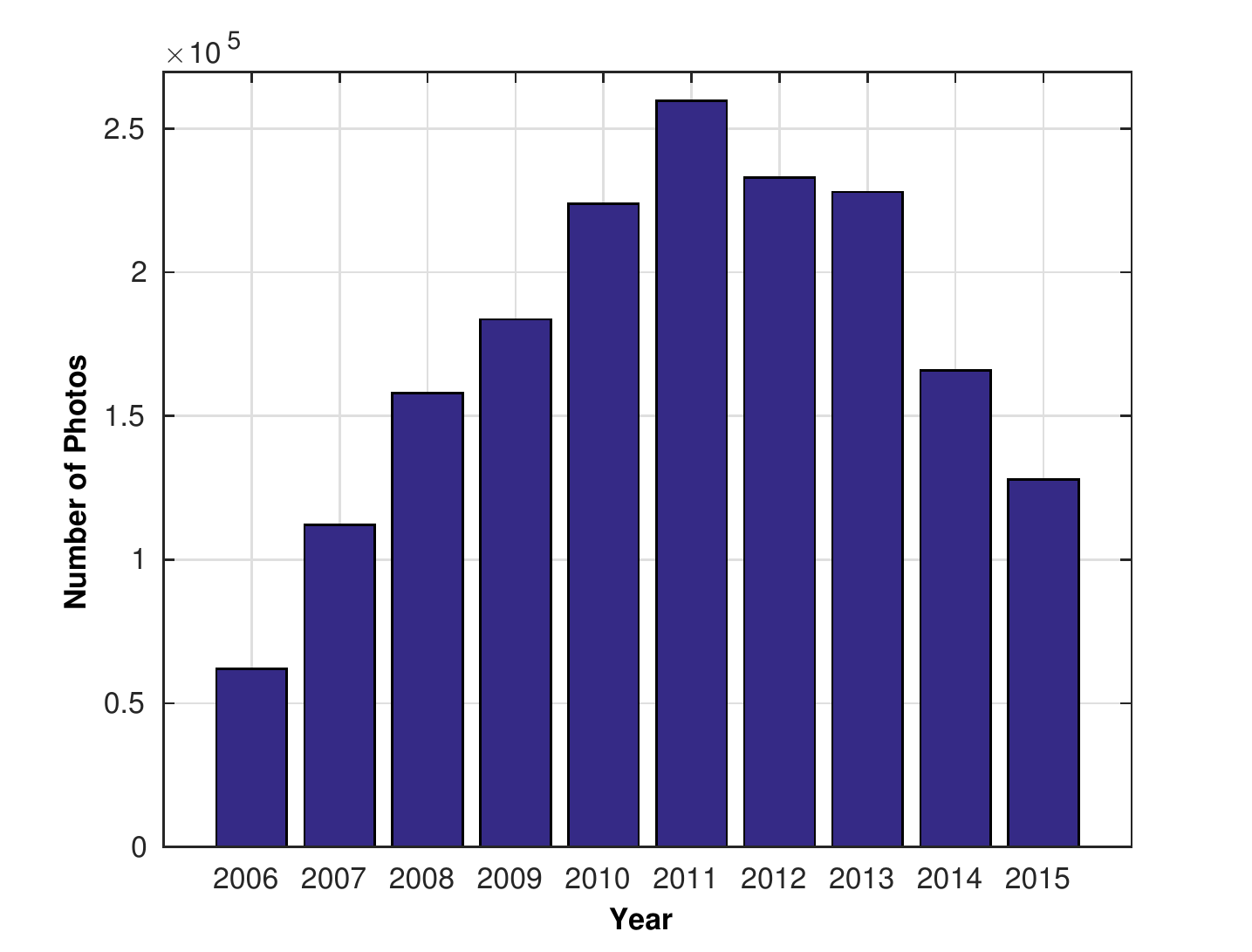}\label{fig:numPhotosPerYear}}\hspace{0pt}
	\subfigure{\includegraphics[width=0.49\linewidth,trim=0 0 0 0,clip]{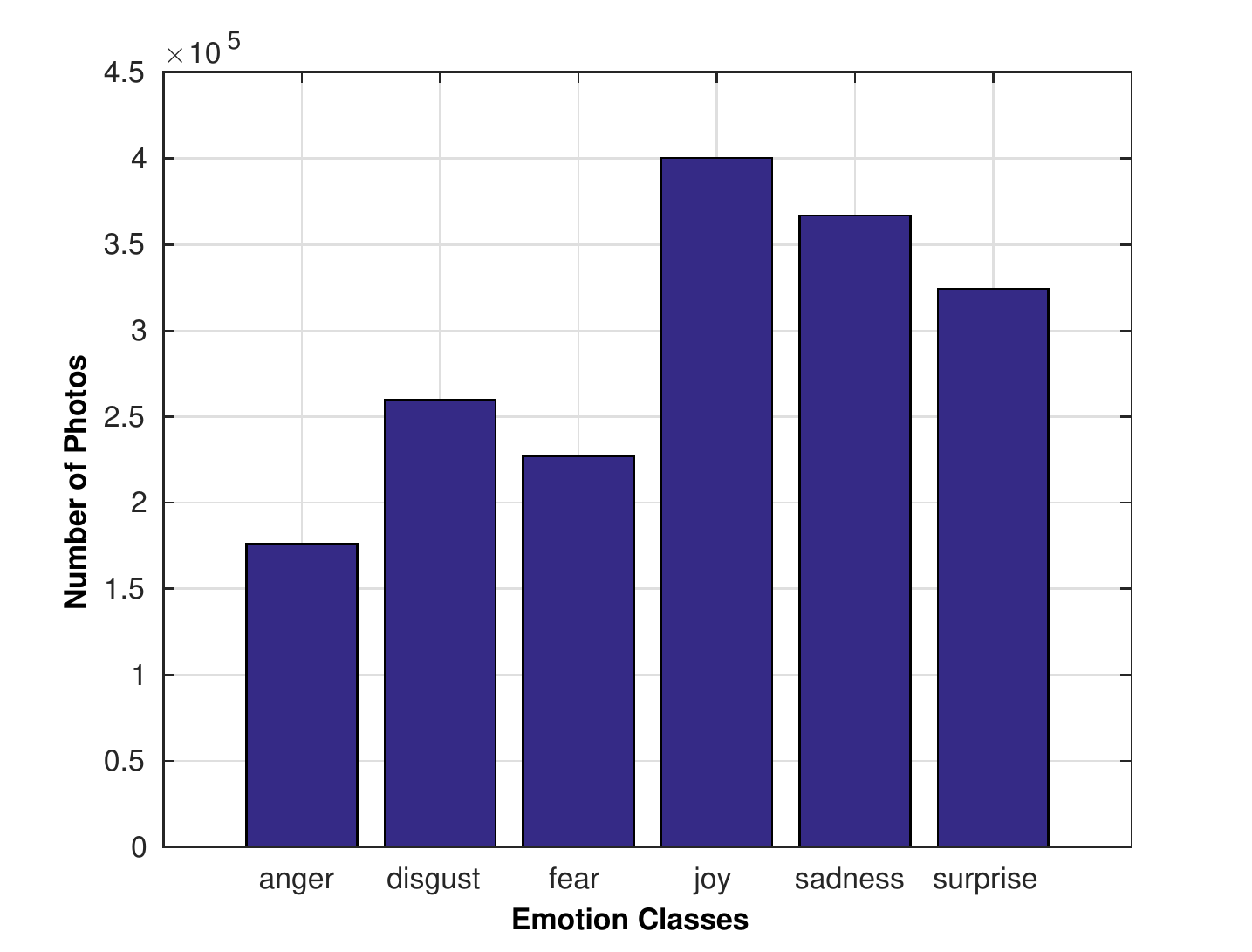}\label{fig:numPhotosPerEmotion}}\hspace{0pt}
	\vspace{-6ex}
	\caption{Number of geo-referenced photos in San Francisco area. Left: per year; right: per emotion. }
	\label{fig:distribution}
	\vspace{-2ex}
\end{figure}

\noindent \textbf{Geotagged Photos}
We download geotagged photos from Flickr for San Francisco city for the ten year period from 2006 to 2015. These are the images we label with our emotion classifier. The total number of images is around $1.9$ million. However, some of the images are too dark/light, too small, or just a placeholder in Flickr, and so we perform a simple filtering step to remove these images. The dataset after filtering contains 1,753,903 images. The distribution by year and predicted emotion can be seen in figure \ref{fig:distribution}. 

Figure \ref{fig:numPhotosPerYear} conveys a sense of popularity of the Flickr platform over the ten year period. The number of uploaded photos reaches a peak of 259,741 in 2011 and then falls each year after that. This decline is interesting although we leave it to the reader to stipulate on its cause. Figure \ref{fig:numPhotosPerEmotion} shows the distribution of emotions as predicted by our classifier. There are more joy and sadness images than other emotions.
% instgram initial release 2010 and become very popular in 2012, active user 100 million and 300 million in 2014.

\vspace{-1ex}
\subsection{Spatial Hotspot Analysis}
\label{sec:spatialresult}
We now present the results of our spatial hotspot detection. 
%Recall that after applying our emotion classifier to each of the Flickr images, we have one $1000 \times 1000$ grid for each emotion where the bin values represent the ratio of images from 2006 to 2015 labeled with that emotion to all the images in that bin. We then compute the Gi* statistic for each bin that has at least one image (of any emotion) to obtain the statistically significant hot and cold clusters.
We use the optimized hotspot analysis tool\footnote{http://desktop.arcgis.com/en/arcmap/10.3/tools/spatial-statistics-toolbox/optimized-hot-spot-analysis.htm} in ArcGIS to compute and visualize our results. 
%The hotspot regions corresponding to emotion disgust and joy can be seen in Fig. \ref{fig:spatialHotspotDection}. 
%Red indicates statistically significant hot spots, while blue represents statistically significant cold spots. The intensity of the color corresponds to the confidence score of the decision, i.e. the reddest areas indicate a $99\%$ confidence level.

One of the challenges of our work is that there is no ground-truth for evaluation.
%quantitative analysis. 
Nonetheless, we make the following qualitative observations from the results in Fig. \ref{fig:spatialHotspotDection}: 
\textbf{(i)} Different emotions have distinct spatial patterns (we only visualize the results of emotions joy and disgust for illustration). This indicates that our emotion classifier is detecting consistent signals in the geotagged photos.
\textbf{(ii)} The detected hotspots make sense. For example, Fig. \ref{fig:joy} shows that joy hotspots are detected at the San Francisco botanical garden, Alamo square park, AT$\&$T park, Candlestick park, and the Fort Mason chapel. Further, these locations are detected as coldspots or not being significant for the other emotions.
%\item Some places are considered hotspots for multiple emotions. For example, the beaches and cliffs along the west coast of San Francisco are considered hotspots for both anger and surprise. 
\textbf{(iii)} Some places are a mix of all emotions. These places, such as downtown San Francisco, have a wide variety of scenes which results in a relatively balanced distribution. 
%However, this balanced ratio is statistically lower than the global mean (in equation \ref{eq:Gi-more}) of any one emotion and so these locations actually show up as cold spots for all emotions.  

%In addition to hotspot detection from geostatistics, we also tried to identify interesting spatial patterns in the mapped emotions using standard clustering methods from machine learning. We chose mean shift clustering \cite{mean_shift_PAMI2002} since we have no a priori knowledge about the correct number of clusters.
%Mean shift is a non-parametric, iterative feature subspace analysis technique for detecting the modes of a density function. We use a Gaussian kernel, and run the algorithm multiple times with different values of the bandwidth parameter to produce results with differing numbers of clusters. We observe, however, that the clustering partitions San Francisco spatially based on the distribution of an emotion (as one would expect it to). This partitioning is not useful in determining where concentrations of an emotion are and is very sensitive to spatial outliers. The Gi* statistic is a more suitable method for locating these concentrations.

\begin{figure*}[!ht]
	\vspace{-1ex}
	\centering
	\subfigure{\includegraphics[width=0.3\linewidth,trim=0 0 0 0,clip]{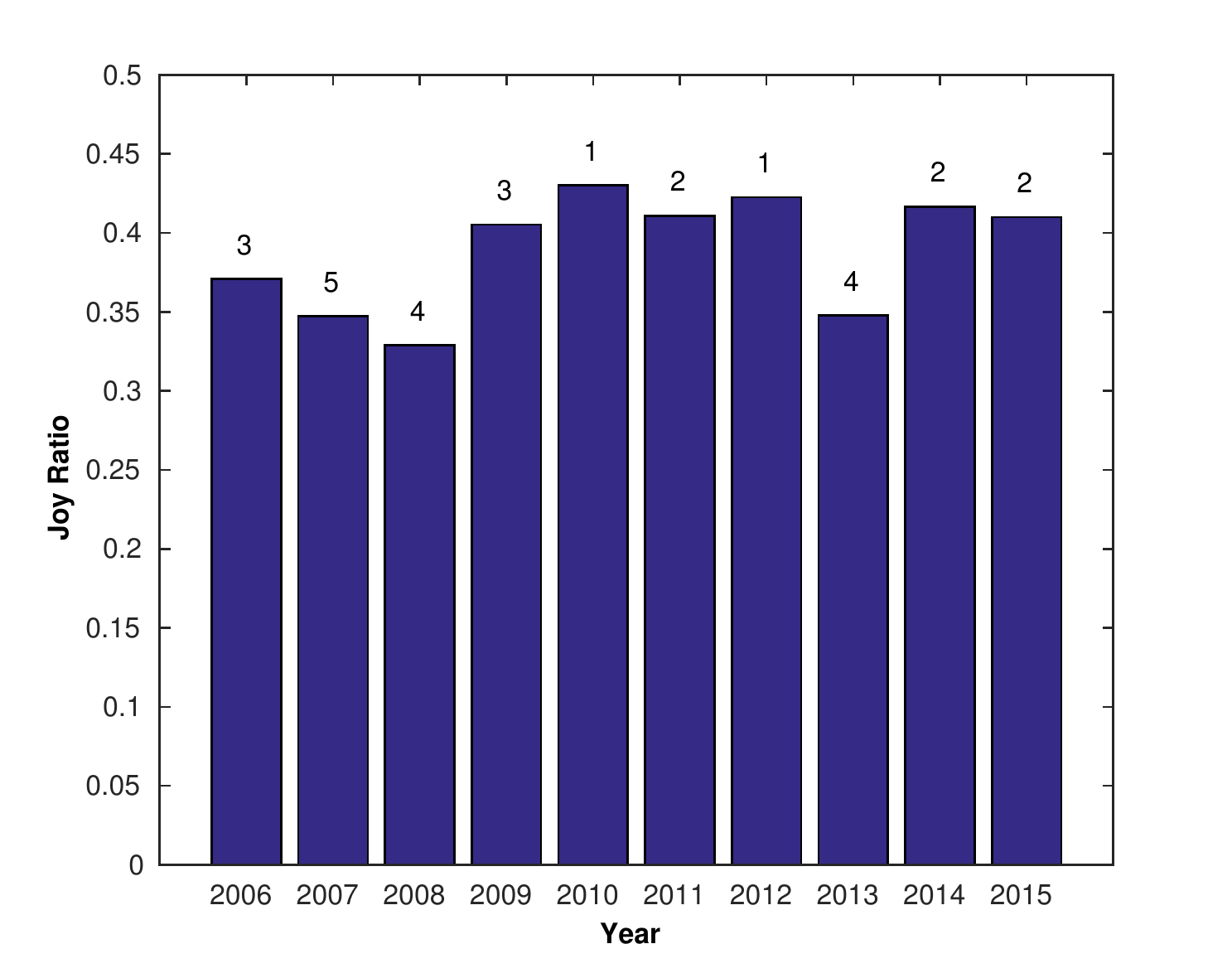}\label{fig:att}}\hspace{40pt}
	\subfigure{\includegraphics[width=0.3\linewidth, height=120pt, trim=0 0 0 0,clip]{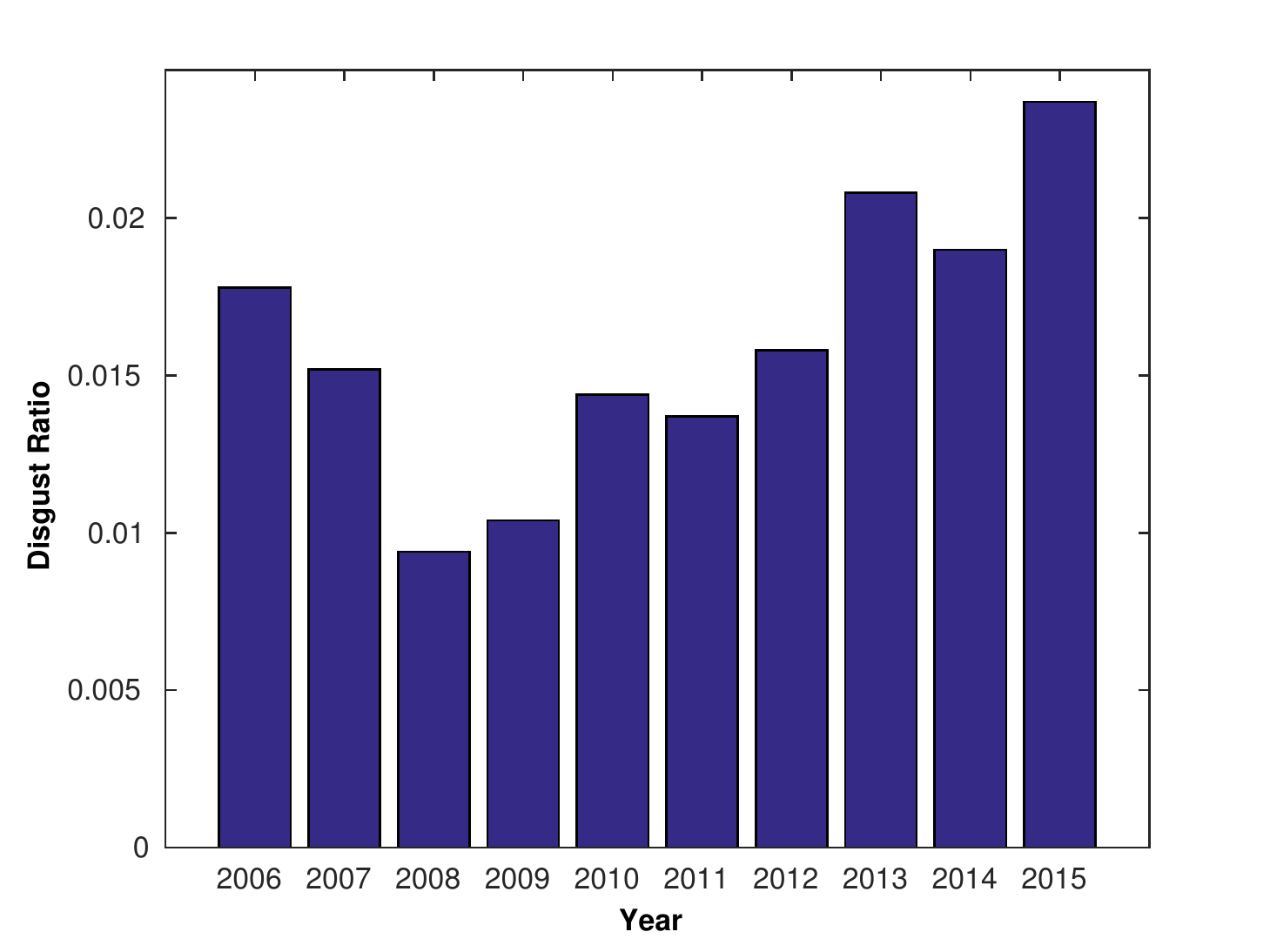}\label{fig:mission}}\hspace{0pt}
	\subfigure{\includegraphics[width=0.3\linewidth, height=120pt, trim=0 0 0 0,clip]{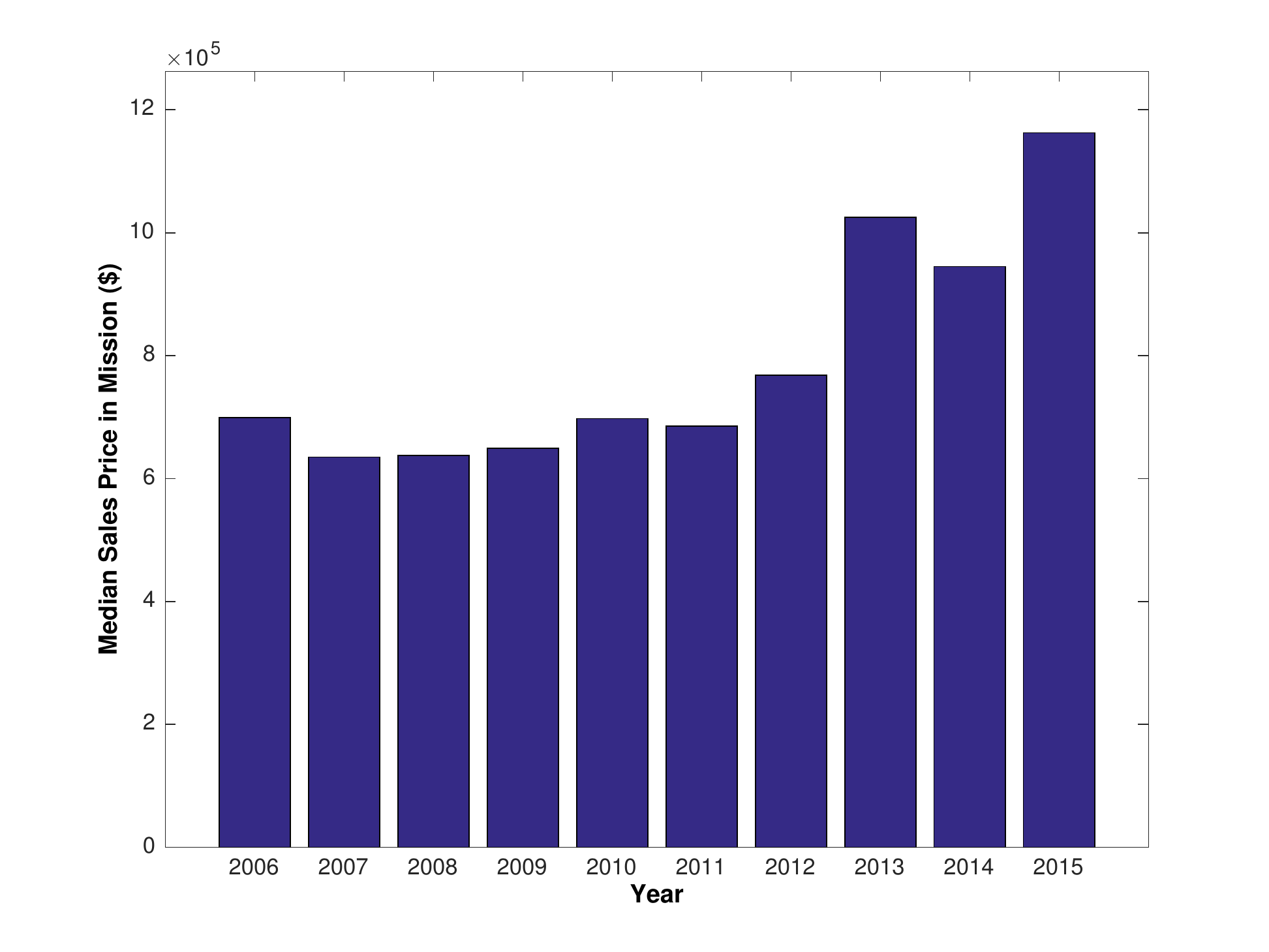}\label{fig:missionPriceTrend}}\hspace{0pt}
	\vspace{-4ex}
	\caption{Observed temporal trends for select regions. Left: Joy ratio computed yearly for AT$\&$T park. Shown above the bars are the end of season rankings of the SF Giants who play at the park. Notice the strong correlation. Middle: Disgust ratio computed yearly for the Mission neighborhood. Notice the steady increase since 2008 which is about when it started to become very popular with young professionals in the tech industry. Right: The average house price in the Mission neighborhood from a real estate website\textsuperscript{\ref{note}}. Notice the correlation with the disgust ratio.}
	\label{fig:attandmission}
	\vspace{-3ex}
\end{figure*}

%\FloatBarrier
\vspace{-1ex}
\subsection{Spatio-Temporal Hotspot Detection}
\label{sec:spatio-temporalresult}
%We now present the results of our spatio-temporal hotspot detection for 2006-2015. 
\noindent The goal here is to identify locations that are significiant in both space and time. We first conduct global detection and then perform temporal analysis for select locations.
%We use the emerging hotspot analysis tool\footnote{https://desktop.arcgis.com/en/arcmap/latest/tools/space-time-pattern-mining-toolbox/emerginghotspots.htm} in ArcGIS to visualize new, consecutive, intensifying, persistent, diminishing, sporadic, oscillating and historical hotspots. 

%\begin{figure}%[htb]
%	\centering
%	\includegraphics[width=1.0\linewidth,trim=0 20 0 0,clip]{figures/space_time.eps}
%	\vspace{-2ex}
%	\caption{The spatio-temporal hotspots detected in San Francisco for the period 2006-2015. This is for the emotion joy. This figure is best viewed in color. \vspace{-4ex}}
%	\label{fig:spatialTemporalHotspotDection}
%\end{figure}
\noindent \textbf{Global Detection}
Fig. \ref{fig:spatialTemporalHotspotDection} shows the spatio-temporal hotspots detected for the north-east part of San Francisco. 
We make the following observations: 
\textbf{(i)} Pier 70 (blue box) is detected as a new hotspot which means that is a statistically significant hot spot for the final time step (2015) but has never been a statistically significant hot spot before.
%\footnote{https://desktop.arcgis.com/en/arcmap/latest/tools/space-time-pattern-mining-toolbox/learnmoreemerging.htm}. 
This makes sense since the Pier 70 buildings were recently renovated in 2014 to host large corporate parties, concert events, expositions, etc.
\textbf{(ii)} Tourist destinations and public spaces are detected as intensifying hotspots which means they have been a statistically significant hot spot for ninety percent of the time-step intervals, including the final time step, and the intensity of clustering of high counts in each time step is increasing overall and that increase is statistically significant. The fact that these are intensifying and not just persistent hotspots is interesting. We postulate that it is due to the economic recovery that has occurred during the latter part of our time period which especially affects the tourist and leisure industry.
\textbf{(iii)} Many locations are detected as consecutive hotspots which means there is a single uninterrupted run of statistically significant hot spot bins in the final time-step intervals but the location has never been a statistically significant hot spot prior to the final hot spot run and less than ninety percent of all time-steps are statistically significant hot spots. These locations also tend to be detected as cold spots or as having no significance in the spatial hopspot results in Fig. \ref{fig:joy}. Taken together, these results indicate that these locations have recently become hotspots. This again could be the result of the improved economy. It could also be the result of more photos being captured with GPS-enabled smartphones recently and thus having more accurate location information. This would make the photos more concentrated. 
%This requires further investigation.

\noindent \textbf{Local Analysis}
%We here perform more detailed year-to-year analysis at select locations to better understand the temporal trends.
AT$\&$T park shows up as a spatio-temporal hotspot with respect to joy. It is also a location whose sentiment one might expect to be correlated with the performance of the professional baseball team that plays there, the SF Giants. To investigate this, we calculate the joy ratio per year for a window centered on the park. These values are plotted in Fig. \ref{fig:att}. Shown above each bar is the end-of-season ranking of the Giants for each year. The joy ratio and ranking are clearly correlated demonstrating that we are able to detect public sentiment from geotagged photos.

We also perform this local temporal analysis for another location, the Mission, for the emotion disgust. The Mission is one of the less expensive residential neighborhoods in San Francisco and is shown to exhibit a relatively large number of disgust spatial hotspots as shown Fig. \ref{fig:disgust} (this figure also delineates the neighborhood). We compute the per year disgust ratio in a window centered on the Mission and plot the results in Fig. \ref{fig:mission}. There is a clear increasing trend since 2008 which is about when the Mission started to become very popular with young professionals in the tech industry. These were not the traditional Mission residents and the detected increase in disgust could be a result of their reaction to the dirtiness, etc. of the streets. In fact, the yearly disgust ratio is strongly correlated with the average home price for the Mission, shown in Fig. \ref{fig:missionPriceTrend}\footnote{\label{note}http://www.trulia.com/real\_estate/Mission-San\_Francisco/1436/market-trends/}. This increase in housing prices is likely also a result of the new demographic.

\vspace{-2ex}
\section{Conclusions}
\label{sec:conclusion}
We conduct the first investigation into using geotagged social multimedia for spatio-temporal sentiment hotspot detection. We leverage deep ConvNets to develop an emotion classifier to predict the emotions conveyed in geotagged photos. This allows us to associate sentiment with place. We apply the Getis-Ord Gi* statistic to detect spatial hotspots, and show that different emotions have distinct spatial distributions that match expectations. We detect emerging concentrations of emotions through spatio-temporal hotspot detection and show that year-by-year analyses of select locations are correlated with known events.

%The problem of spatio-temporal sentiment analysis using geo-referenced social multimedia is a challenging but rich research problem. There are many interesting directions to extend our work including novel ConvNets architectures for emotion recognition in images, applying the framework to geotagged videos, and multi-class comparative spatio-temporal hotspot detection.
\vspace{-2ex}
\section{Acknowledgments}
We gratefully acknowledge the support of NVIDIA Corporation through the donation of the Titan X GPU used in this work. This work was funded in part by a National Science Foundation CAREER grant, \#IIS-1150115, and a seed grant from the Center for Information
Technology in the Interest of Society (CITRIS). We would like to thank the UC Merced Spatial Analysis and Research Center (SpARC) for help with the hotspot analysis.
\vspace{-2ex}
\bibliographystyle{abbrv}
\bibliography{sigproc}  % sigproc.bib is the name of the Bibliography in this case

\begin{thebibliography}{10}

\bibitem{CrandallMapWorld09}
D.~Crandall et~al.
\newblock {Mapping the World's Photos}.
\newblock In {\em WWW}, 2009.

\bibitem{imagenet_cvpr09}
J.~Deng et~al.
\newblock {ImageNet: A Large-Scale Hierarchical Image Database}.
\newblock In {\em CVPR}, 2009.

\bibitem{ekman_6emotions}
P.~Ekman et~al.
\newblock {What Emotion Categories or Dimensions can Observers Judge from
  Facial Behavior}.
\newblock {\em Emotion in the Human Face}, 1982.

\bibitem{Haysim2gps08}
J.~Hays and A.~A. Efros.
\newblock {IM2GPS: Estimating Geographic Information from a Single Image}.
\newblock In {\em CVPR}, 2008.

\bibitem{emotion_video_jiang_2014}
Y.-G. Jiang et~al.
\newblock {Speech Emotion Recognition Using Deep Neural Network and Extreme
  Learning Machine}.
\newblock In {\em AAAI}, 2014.

\bibitem{AlexImageNet12}
A.~Krizhevsky et~al.
\newblock {ImageNet Classification with Deep Convolutional Neural Networks}.
\newblock In {\em NIPS}, 2012.

\bibitem{shape_emotion_mm12_Lu}
X.~Lu et~al.
\newblock {On Shape and the Computability of Emotions}.
\newblock In {\em ACM MM}, 2012.

\bibitem{mann_kendall_test_1945}
H.~B. Mann.
\newblock {Nonparametric Tests Against Trend}.
\newblock {\em Econometrica}, 1945.

\bibitem{ord_getis_1995}
J.~Ord and A.~Getis.
\newblock {Local Spatial Autocorrelation Statistics: Distributional Issues and
  an Application}.
\newblock {\em Geographical Analysis}, 1995.

\bibitem{smartUrban15}
S.~Paldino et~al.
\newblock {Urban Magnetism Through The Lens of Geo-tagged Photography}.
\newblock {\em arXiv preprint arXiv:1503.05502}, 2015.

\bibitem{emotion_image_Peng_2015}
K.-C. Peng et~al.
\newblock {A Mixed Bag of Emotions: Model, Predict, and Transfer Emotion
  Distributions}.
\newblock In {\em CVPR}, 2015.

\bibitem{emotionTwitter_2016}
B.~Resch et~al.
\newblock {Citizen-Centric Urban Planning through Extracting Emotion
  Information from Twitter in an Interdisciplinary Space-Time-Linguistics
  Algorithm}.
\newblock {\em Urban Planning}, 2016.

\bibitem{3dModelSnavely08}
N.~Snavely et~al.
\newblock {Modeling the World from Internet Photo Collections}.
\newblock {\em IJCV}, 2008.

\bibitem{Emotion_Large_AAAI16_You}
Q.~You et~al.
\newblock {Building a Large Scale Dataset for Image Emotion Recognition: The
  Fine Print and The Benchmark}.
\newblock In {\em ACM MM}, 2016.

\bibitem{landuse_yi_sigspatial_2015}
Y.~Zhu and S.~Newsam.
\newblock {Land Use Classification Using Convolutional Neural Networks Applied
  to Ground-Level Images}.
\newblock In {\em ACM SIGSPATIAL}, 2015.

\end{thebibliography}
% ACM needs 'a single self-contained file'!
%\balancecolumns
\end{document}